# Deep Learning for Forecasting the Energy Consumption in Public Buildings

Viorica Rozina Chifu, Cristina Bianca Pop, Emil St. Chifu, Horatiu Barleanu
Department of Computer Science
Technical University of Cluj-Napoca
Cluj-Napoca, Romania
{viorica.chifu, cristina.pop, emil.chifu}@cs.utcluj.ro

*Abstract*—In this paper we propose a Long Short-Term Memory Network based method to forecast the energy consumption in public buildings, based on past measurements. Our approach consists of three main steps: data processing step, training and validation step, and finally the forecasting step. We tested our method on a data set consisting of measurements taken every half an hour from the main building of the National Archives of the United Kingdom, in Kew and as evaluation metrics we have used Mean Absolute Error (MAE) and Mean Absolute Percentage Error (MAPE).

## I. INTRODUCTION

The problem of forecasting the electrical energy consumption becomes more and more important as the worldwide energy demand is increasing. According to a study provided by [17], at worldwide, between 2018 and 2050, there will be an increase of 50% in energy consumption. The biggest increases in energy demand comes from countries with strong economic growth, such as the Asian countries. Unlike other energy sources, such as coal, oil and natural gases, the electric energy is hard to store at a large scale and is expensive to store at a small scale. In an ideal situation, the power plants produce the exact amount of energy that is needed. However, ideal situations do not exist in practice; for a power plant to run as economically as possible, it would be highly beneficial to know the electric load in advance. By possessing this forecast information the working parameters of the plant can be finely tuned and the supply of primary energy sources (gas, oil, coal etc.) can be bought and stocked in a more economical manner. The energy consumers would also benefit from a precise forecast.

In this paper we propose a neural network, namely Long Short Term Memory to forecast the energy consumption in public buildings by using past measurements. Our method consists of three main steps: data processing, neural network training and validation, and finally a forecasting step. We validated our method on a data set provided by [18] consisting of measurements taken every half an hour from the main building of the National Archives of the United Kingdom, in Kew. We measured the prediction accuracy by using the mean absolute error and the mean absolute percentage error.

The paper is structured as follows: section 2 overviews related work; section 3 presents the deep learning based method for forecasting the energy consumption; section 4 discusses experimental results; section 5 presents conclusions.

## II. RELATED WORK

In the field of forecasting the energy consumption, most of the approaches in the research literature use machine learning techniques to predict the energy consumption [1] - [14].

In [1], Random Forests and Feed-Forward Back-Propagation (FFBP) are used for predicting the energy consumption at every hour of the day by a HVAC system in a hotel. The data used in the experiments was collected during a year and contains information about the energy consumption, the weather parameters, as well as the occupancy rate of the hotel. Additional features, such as hour of the day, day of the week, and month of the year, were extracted from the energy consumption historical data. The experimental results conclude that the FFBP performs slightly better than the Random Forest.

[2] present a system that integrates three neural networks, namely a Fully Connected Neural Network, a CNN, and a LSTM network, to forecast the monthly energy consumption. The experiments are performed on real data collected during two years by a power company. This data set provides information about the monthly energy consumption of residential, business, and industrial customers in Brazil. The prediction is made on a month, by using the energy consumption in the previous 12 months in order to predict the next month. A sliding window technique is used to generate samples for the 13 months. As for the evaluation of the obtained results, the metrics used are the mean absolute error (MAE), the mean absolute percentage error (MAPE) and the median absolute percentage error (MdAPE). The results obtained with the three neural networks were compared also to the procedure currently used by the power company, which involves considering the average of a customer's consumption over the last months as the predicted value for the next one. All three neural networks integrated in the system provide more accurate results compared to the procedure previously mentioned. Out of the three neural networks used, the one producing the best results is the LSTM.

[4] describe two stochastic machine learning methods, namely the Conditional Restricted Boltzmann Machine (CRBM) and the Factored Conditional Restricted Boltzmann Machine (FCRBM) for predicting the energy consumption. The two methods are tested on a data set consisting of values of energy consumption and some other electrical amounts, measured for a household over a period of 4 years, at intervals of one minute. The data corresponding to the first three years are used for training, while the fourth year is used for testing. The experiments use different forecast horizons and various resolutions. For evaluating the performance, the results obtained are compared the Artificial Neural Networks, SVM and Recurrent Neural Networks. The model which performed best was FCRBM, surpassing the other state-of-the-art forecasting methods.

[3] propose a method based on linear regression to analyze the energy consumption in Germany. They investigate the energy consumption in the industrial, residential, and transportation sectors. With the help of some forecasting tools, the authors use the best regression model in order to predict the energy consumption. The predictors used are: per year GDP, population growth and industrial growth rate.

In [5], a regression analysis method based on multiple linear regression is implemented and utilized for predicting the energy consumption of a supermarket. The authors use the real consumption and weather data for the development of the regression method. The energy consumption will be estimated based on the outdoor temperature and the outdoor air humidity. The method predicts the energy for the following years. Each of these parameters can have its own linear relationship with the year that the prediction is made for.

In [16] a decentralized solution based on blockchain for constructing and managing Virtual Power Plants of small-scale prosumers is proposed. The VPP is constructed in a hierarchical manner in which the VPPs on the higher levels are constructed from the smaller VPPs or prosumers from the lower layers. The nodes of the blockchain network are represented by prosumers, while the transactions are represented by the monitored energy values of the prosumers. For validation a proof of concept prototype using Ethereum was implemented.

## III. LSTM based Method for Forecasting the Energy Consumption

The method we propose uses the Long Short-Term Memory Network (LSTM) [15] for forecasting the energy consumption in public buildings and consists of the steps described below.

**Data Processing**. The data processing involves the following steps: (*i*) the data set is parsed, the energy consumption measurements are saved into a matrix and the timestamps into an array; (*ii*) the timestamps are converted into date and time objects; (*iii*) two new matrices are built, one containing the neural network input as multidimensional input feature vectors, and the other consisting of energy consumption measurements used for error evaluation.

The features of the input feature vectors are the same as the features of the measurements vectors used in error evaluation. During our experimental evaluation, we tested with different features for these vectors, until arriving at the most accurate results.

In our first experiment we separate the timestamp into four values: year, month, day, and half hour. The input data has now the form of an array of four dimensional vectors, each dimension representing one of the aforementioned values:

$$\overrightarrow{input_1} = < year, month, day, half\ hour >$$
$$\vdots$$
$$\overrightarrow{input_m} = < year, month, day, half\ hour >$$

where $m$ represents the number of examples in the data set.

In our second experiment, we enriched the input vectors with as much seasonal data as possible. Besides the features present in the previous step, we also used some additional input features: day of the week, day of year, week of the year.

In order to further improve the prediction accuracy, in our third experiment we combined a regression based approach with a time series forecasting method. The idea is to add measurements from previous half hours as input features, in a sliding window manner. This approach has one disadvantage: when forecasting, the network can only predict one half hour consumption at a time, which has to be introduced as an input feature for the next half hour. For a short forecasting horizon, the prediction is very precise. However, for a large horizon, the error propagates as we get far away from the end of the training interval. And consequently, the predicted values as well will become less and less accurate for longer-term forecasts. For this approach, the format of an input vector is the following:

$$\overrightarrow{input_m} = < h_h, d_m, d_w, w, d_y, y, sl_1, sl_2, \cdots, sl_n >$$

where $sl_1$ to $sl_n$ comprise the sliding window, and $n$ is the sliding window size.

**Neural Network Training.** For the training and validation cycle, the data set is divided into two subsets, one for training and one for validation. For a given number of epochs, the network is trained and validated. A training phase involves the following steps. First, the training subset is partitioned into batches, and then, for each batch, there is a forward and a back-propagation pass through the network. In the forward pass, the batch is fed to the network, and the result output constitutes the predicted consumption values. This predicted consumption is then compared with the actual measurements, to give the Mean Absolute Error (MAE). In the back-propagation pass, the Adam optimization algorithm adjusts the weights of the network in order to minimize the errors. Finally, the average of the errors for each batch is computed in order to obtain the error on the training epoch, and then the trained neural network model is saved.

Since we use a sliding window method for solving the energy forecasting as a time series forecasting problem, the data subset used for validation has to start at the end of the training subset. So, several measurements (depending on the size of the sliding window) from the training subset will be included as input values for several validation examples. The validation step takes each input example in turn (the examples being not separated into batches). The validation example is fed to the network in order to obtain a predicted value, which is then normalized and included in the input data of the next input example. Then the Mean Absolute Error (MAE) and the Mean Absolute Percentage Error (MAPE) are computed. Finally, the overall error value on the whole validation subset is computed for the two above-mentioned error metrics.

**Forecasting the Energy Consumption.** In this step, an already trained model is used in order to forecast the energy consumption on the testing data set. The parameters (i.e. the weights) and the state of the network after a certain epoch are used for making predictions. The forecasting can be made on a given horizon, where the starting point of the horizon must be the first validation example after the last training example. Having this feature is extremely useful. We can use a part of the validation subset, which coincides with the chosen forecasting horizon. Having the trained neural network model, we can choose different horizons and notice how the error changes depending on the lengths of the horizons.

**Evaluating the Results.** We evaluated our experimental results by using different MAE and the MAPE metrics. The graphs generated during the training and validation process represent (a) the evolution of the Mean Absolute Error (MAE) for the training step of each epoch, (b) the evolution of the Mean Absolute Error (MAE) and the Mean Absolute Percentage Error (MAPE) for the validation step of each epoch, and (c) the actual measurements of the energy consumption and the values of the predicted consumption.

## IV. EXPERIMENTAL RESULTS

We evaluated our method on a dataset [18] consisting of measurements taken every half an hour from the main building of the National Archives of the United Kingdom, in Kew. In the dataset, each row represents a day of measurements and each column represents the energy consumption for a given half an hour.

We detail, in what follows, the evolution of the values of the error metrics when using LSTM. The evolution is tracked according to different sliding window sizes (where the size of a sliding window is the number of half hours in the window). The sliding window size is increased incrementally with one more half an hour for the different experiments.

For the LSTM, we achieved the best results for the following values of the hyperparameters: (i) batch size = 10, (ii) LSTM number of layers = 3, (iii) sliding window size = 480, (iv) number of LSTM units per layer = 256. The training and validation losses (i.e. the Mean Absolute Error) as tracked during the training process are shown in Figure 1. The measured energy consumption and the predicted consumption for ten days (i.e. for 480 half hours) are displayed in Figure 2. The predicted energy consumption as tracked in the figure is obtained by a network state in the epoch that had the minimum validation errors. The next step is to see how much the sliding window size could be increased until the errors stop decreasing or even start to increase. Besides the varying size of the sliding window, we keep the same values for the hyperparameters as in the first experimental run, just illustrated above. Figures 3-6 present the minimum validation loss for the following varying sizes of the sliding window: 960, 1920, 3840, and 4320 half hours, respectively (i.e. 20, 40, 80, and 90 days, respectively). It can be noticed that the minimal errors are obtained when using a window size of 3840 previous measurements (on a half an hour basis). However the error does not vary greatly between the window sizes of 1920, 3840, and 4320. Another aspect is that the algorithm becomes more and more computationally intensive as the sliding window size increases. Consequently, for the next experimental tuning of the parameters, we will use a rather reduced window size, i.e. between 1920 and 3840, as a subrange of the best range of window sizes noticed so far, 1920-4320. Figure 7 illustrates the measured and predicted consumption for a window size of 3840 half hours (which is the best window size according to our experiments).

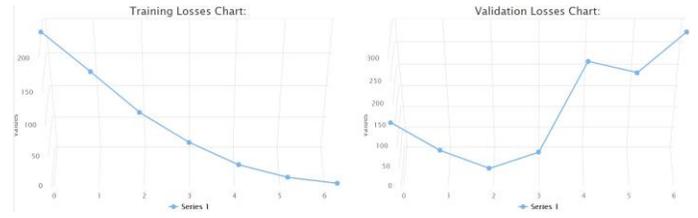

Fig. 1. Sliding window size of 480 half hours

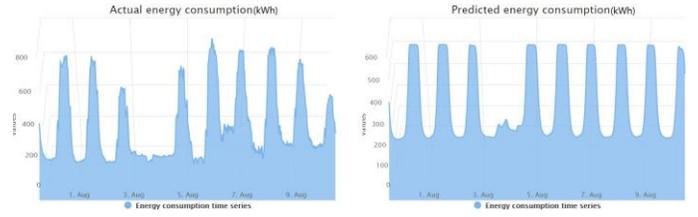

Fig. 2. Measured consumption and predicted consumption when using a sliding window size of 480 half hours

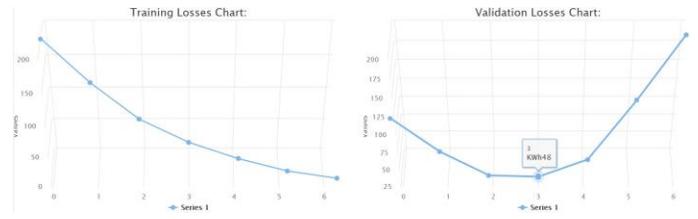

Fig. 3. Sliding window size: 960

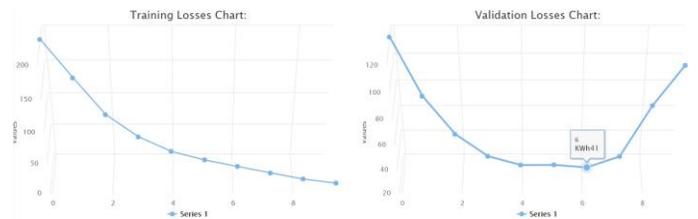

Fig. 4. Sliding window size: 1920

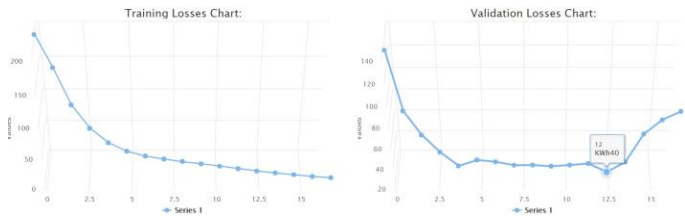

Fig. 5. Sliding window size: 3840

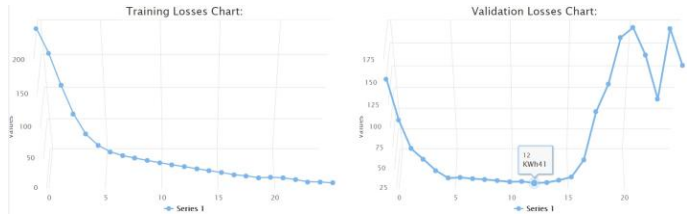

Fig. 6. Sliding window size: 4320

**Dropout.** When we added dropout layers between the LSTM layers the result was a reduction of the mean absolute errors by about 25 percent, i.e. from 40 kWh to 32 kWh. When adding dropout layers in the experiment with the sliding window size of 3840 half hours (which is our best experiment), and setting the dropout keep probability to 0.5, the result training and validation losses are the ones shown in Figure 8. The measured and predicted energy consumption for this new experiment are displayed in Figure 9.

**Tuning the Size of the Forecasting Horizon.** We now track the evolution of the values of the error metrics as the forecasting horizon increases. The hyperparameters of the network and the characteristics of the dataset for the experiments detailed below are: batch size = 10, dropout rate = 0.5, learning rate = 1.0E-4, window size = 3840, training data set length = 100000, total data set length = 131520, training data start = 15000. We computed the validation error during the training process by using a 10 days forecasting horizon (see Figure 10).

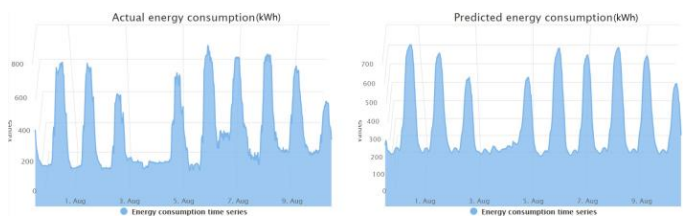

Fig. 7. Actual and predicted consumption for a window size of 3840 half hours

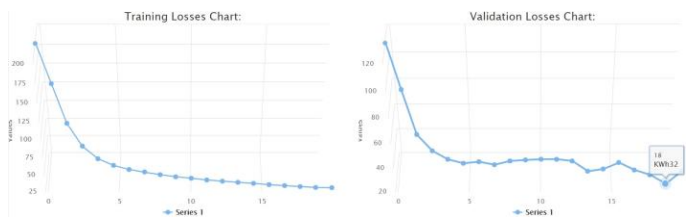

Fig. 8. Training and validation losses with 0.5 dropout probability for 3840 half hours window size

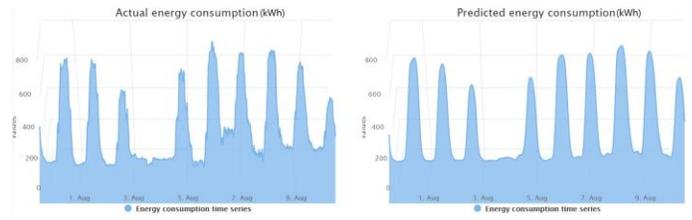

Fig. 9. Measured and predicted consumption using dropout for 3840 half hours window size

Figures 10 - 15 show the values of the predicted consumption when using a 10, 20, 30, 40, 50, and 100 days forecasting horizon respectively. It can be noticed that both the MAE and MAPE increase as the forecasting horizon becomes much longer. However, an interesting aspect is that the error is lower for the 20 days ahead forecast than the 10 days ahead.

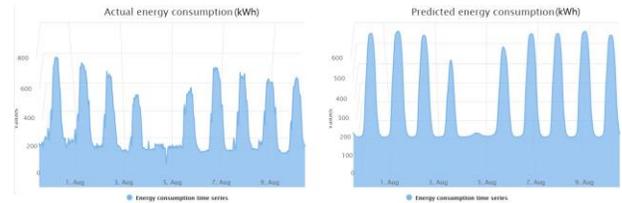

Fig. 10. Ten days forecasting horizon: MAE = 22.56; MAPE = 7.96

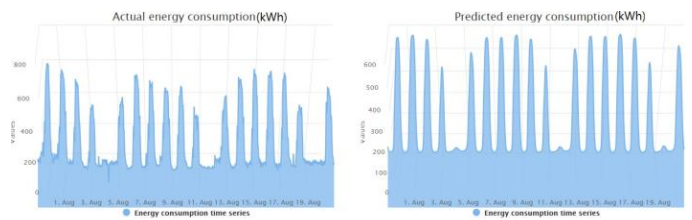

Fig. 11. 20 days forecasting horizon: MAE = 21.32; MAPE = 7.6

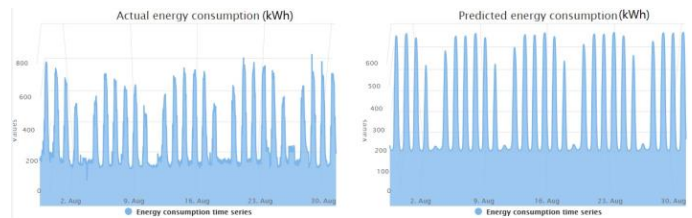

Fig. 12. 30 days forecasting horizon: MAE = 22.97; MAPE = 8.04

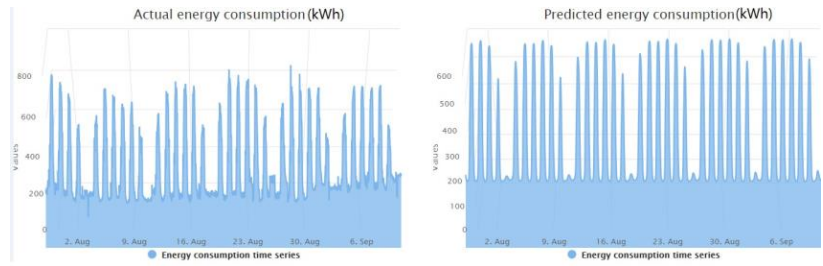

Fig. 13. 40 days forecasting horizon: MAE = 30.15; MAPE = 10.63

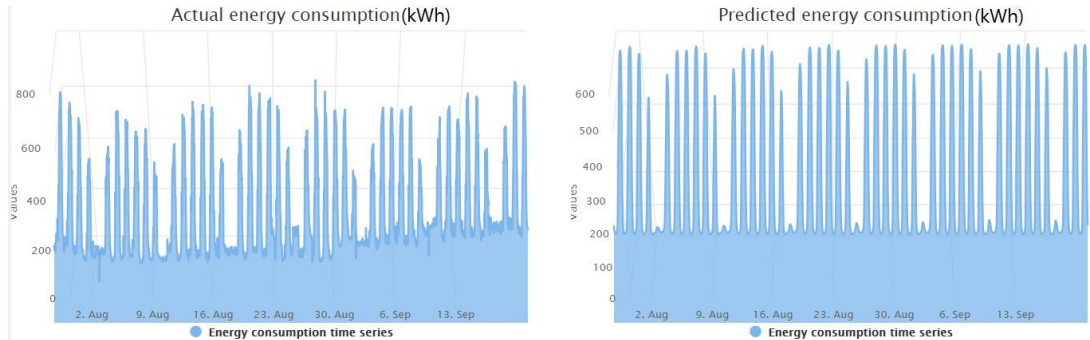

Fig. 14. 50 days forecasting horizon: MAE = 38.27; MAPE = 13.09

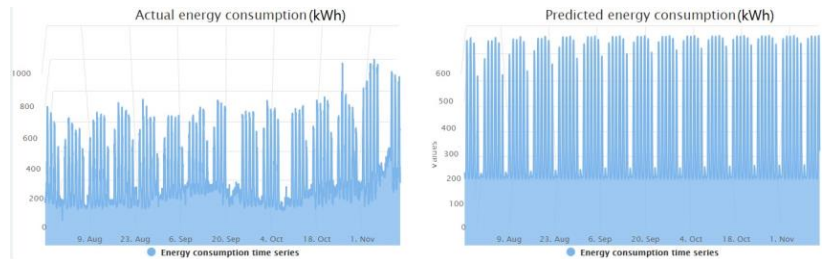

Fig. 15. 100 days forecasting horizon: MAE = 64.54; MAPE = 21.05

## V. CONCLUSIONS

In this paper we presented a deep learning based method that is able to predict the energy consumption in public buildings based on past measurements. The proposed method integrates a neural network, LSTM. We evaluated the neural network on a data set consisting of measurements taken every half an hour from the main building of the National Archives of the United Kingdom, in Kew. As evaluation metrics we have used Mean Absolute Error and Mean Absolute Percentage Error. The obtained results can be used to provide support in the efficient operation of the electricity grid and in the efficient management of energy consumption in public buildings. Also, the prediction of energy consumption could provide valuable information on how energy can be saved.

## ACKNOWLEDGMENT

This work has been conducted within the BRIGHT project grant number 957816 funded by the European Commission as part of the H2020 Framework Programme and it was partially supported by a grant of the Romanian Ministry of Education and Research, CNCS/CCCDI–UEFISCDI, project number PN-III-P3-3.6-H2020-2020-0031.